\documentclass{svproc}

\usepackage{url}

\usepackage{fancyhdr}
\usepackage{amssymb}
\usepackage{mathrsfs} 
\usepackage{amsmath}
\usepackage{color}

\def\bbR{\mathbb{R}}

\def\[#1\]{$$#1$$}

\def\frac#1#2{{#1\over#2}}
\def\eps{\varepsilon}

\usepackage{graphicx}

\begin{document}
\mainmatter              

\title{On the Role of Time in Learning}

\author{Alessandro Betti\inst{1}\inst{2} \and Marco Gori\inst{2}}

\tocauthor{Alessandro Betti, Marco Gori}

\institute{University of Florence, Florence, Italy,\\
\email{alessandro.betti@unifi.it}
\and
SAILab,
University of Siena, Siena, Italy,\\
\email{marco@diism.unisi.it}\\
WWW home page: \texttt{http://sailab.diism.unisi.it}
}

\maketitle              

\begin{abstract}
By and large the process of learning concepts that are
embedded in time is regarded as quite a mature research topic.
Hidden Markov models, recurrent neural networks are, amongst
others, successful approaches to learning from temporal data.
In this paper, we claim that the dominant approach minimizing
appropriate risk functions defined over time by classic 
stochastic gradient might miss  the deep interpretation of
time given in other fields like physics. We show that 
a recent reformulation of learning according to the principle
of Least Cognitive Action is better suited
whenever time is involved in learning.  The principle gives
rise to a learning process that is driven by differential 
equations, that can somehow descrive the process within
the same framework as other laws of nature.
\end{abstract}

\section{Introduction}
The process of learning has been recently formulated under
the framework of laws of nature derived from variational
principle~\cite{DBLP:journals/tcs/BettiG16}.
While the paper addresses some fundamental issues on the
links with mechanics, a major open problem is the one
connected with the satisfaction of the boundary conditions
of the Euler-Lagrange equations of learning. 

This paper springs out from recent studies especially on
the problem of learning visual features
\cite{DBLP:journals/corr/abs-1808-09162,DBLP:journals/corr/abs-1807-06450,DBLP:journals/corr/abs-1801-07110}
and it is also stimulated by a nice analysis on the 
interpretation of Newtonian mechanics equations in the
variational framework~\cite{Stefanelli2013}. 
It is pointed out that the formulation of learning as
Euler-Lagrange (EL) differential equation is remarkably different
with respect to classic gradient flow. 
The difference is mostly originated from the continuous
nature of time; while gradient flow has a truly algorithmic 
flavor, the EL-equations of learning, which are the outcome of
imposing a null variation of the action, can be interpreted as
laws of nature. 

The paper shows that learning is driven by 
fourth-order differential equations that collapses to 
second-order under an intriguing interpretation 
connected with the mentioned result given in~\cite{Stefanelli2013}
concerning the arising of Newtonian laws.



\section{Euler-Lagrange equations}
Consider an integral functional $F\colon X \to \overline\bbR
:=\bbR\cup\{-\infty,+\infty\}$ of the following form 
\begin{equation}
	{\cal A}(q):=\int_{t_1}^{t_N} L(t, q(t), \dot q(t))\, dt
\label{ActionDef}
\end{equation}
where $L\in {\cal C}^1(\bbR\times\bbR^n\times\bbR^n)$ maps a point
$(t,q,p)$ into the real number $L(t,q,p)$
and $t\mapsto q(t)\in \bbR^n$ is a map of $X$. Consider a partition
$t_1<t_2<\cdots<t_N$ of the
interval $[t_1,t_N]$ into $N-1$ subintervals of length $\eps$.
Given a function $q$ one can identify  the point 
$(q(t_1), q(t_2),\dots q(t_N))\in\bbR^N$, and in general one can define
the subset of $\bbR^N$
$$X_\eps:=\{(q(t_1), q(t_2),\dots q(t_N))\in \bbR^N: q\in X\}.$$
Now consider the 
and consider the following ``approximation'' ${\cal A}_\eps\colon
X_\eps\to \bbR$
of the functional integral $F$:
$${\cal A}_\eps(x_1,\dots, x_N):= \eps\sum_{k=1}^{N-1}
L(k, x_k, \Delta_\eps x_k),$$
where $\Delta_\eps x_k=(x_{k+1}-x_k)/\eps$.
The stationarity condition on ${\cal A}_\eps$ is $\nabla {\cal A}_\eps(x)=0$, thus we have
$$\nabla_i {\cal A}_\eps (x)=\eps \nabla_i[L(i-1,x_{i-1},\Delta_\eps x_{i-1})+
L(i,x_i,\Delta_\eps x_i)].$$
Using the fact that $\partial(\Delta_\eps x_i)/\partial x_i=-1/\eps$ and
$\partial(\Delta_\eps x_{i-1})/\partial x_i=1/\eps$ we get
\begin{align}
\nabla_i {\cal A}_\eps (x)&=\eps[L_p(i-1, x_{i-1},\Delta_\eps x_{i-1})\eps^{-1}+
L_q(i,x_i,\Delta_\eps x_i)-L_p(i,x_i,\Delta_\eps x_i)\eps^{-1}]\cr
&=\eps L_q(i,x_i,\Delta_\eps x_i)-\eps{L_p(i,x_i,\Delta_\eps x_i)-
L_p(i-1, x_{i-1},\Delta_\eps x_{i-1})\over\eps}.
\end{align}
This means that the condition $\nabla {\cal A}_\eps(x)=0$ implies
\begin{equation}
	L_q(i,x_i,\Delta_\eps x_i)- 
	\Delta_\eps L_p(i-1, x_{i-1},\Delta_\eps
	x_{i-1})=0,\qquad i=2,\dots, N-1,
\label{EL-disc-eq}
\end{equation}
where, consistently with our previous definition we are assuming that
$\Delta_\eps L_p(i-1, x_{i-1},\Delta_\eps x_{i-1})=[L_p(i,x_i,\Delta_\eps x_i)
-L_p(i-1, x_{i-1},\Delta_\eps x_{i-1})]/\eps$.

This last equation is indeed the discrete counterpart of the
Euler-Lagrange equations in the continuum:
\begin{equation}
	L_q(t, u(t),\dot u(t))-{d\over dt}L_p(t, u(t),\dot u(t))=0,
	\qquad t\in[t_1,t_N].
\label{EL-cont-eq}
\end{equation}
%
%
The discovery of stationary points of the cognitive action 
defined by Eq.~\ref{ActionDef} is somewhat related with
the gradient flow that one might activate to optimize 
${\cal A}$, namely by the classic updating rule
\begin{equation}
	X_{\epsilon} \leftarrow X_{\epsilon} 
	- \eta \nabla {\cal A}_{\epsilon}.
\end{equation}
This flow is clearly different with respect to
Eq.~\ref{EL-disc-eq} (see also 
its continuous counterpart~\ref{EL-cont-eq}).
Basically, while the Euler-Lagrange equations yield an
updating computation model of $x_{i}$, the gradient 
flow moves $X_{\epsilon}$

\section{A surprising link with mechanics}

Let us consider the action
\begin{equation}
	{\cal A} = \int_{0}^{T} dt \ h(t) \bar{L}(x(t),q(t),\dot{q}(t)).
\end{equation}
The Euler-Lagrange equations are
\begin{align}
	h \bar{L}_{q} - \dot{h} \bar{L}_{\dot{q}} - h 
	\frac{d}{dt} \bar{L}_{\dot{q}}=0.
\end{align}
Since $h>0$ we have
\begin{align}
	\frac{d}{dt} \bar{L}_{\dot{q}} +  \frac{\dot{h}}{h}
	\bar{L}_{\dot{q}} - \bar{L}_{q} =0.
\label{GEL-with-h}
\end{align}
In case we make no assumption on the variation then 
these equations must be joined with
the boundary condition $\big[h L_{\dot{q}}\big]_{0}^{T}=0$.
Now suppose $\bar{L}=T + \gamma V$, with $\gamma \in \bbR$.
Then Eq.~\ref{GEL-with-h} becomes
\begin{align}
	\frac{d}{dt} T_{\dot{q}} + \frac{\dot{h}}{h} T_{\dot{q}}
	 - \gamma V_{q} = 0.
\end{align}
The Lagrangian $\bar{L}=T + \gamma V$, with 
$T = \frac{1}{2} m \dot{q}^{2}$ and $\gamma=-1$, and
$h(t) = e^{\theta t}$,
is the one used in mechanics, which returns the Newtonian
equations 
\[
	m \ddot{q} + \theta \dot{q} + V_{q} = 0
\]
of the damping oscillator.
We notice in passing that this equation arises when 
choosing the classic action from mechanics, which 
does not seem to be adequate for machine learning
since the potential (analogous to the loss function)
and the kinetic energy
(analogous to the regularization term) come
with different sign. It is also worth mentioning
that the trivial choice $h=1$ yields a pure oscillation
with no dissipation, which is on the opposite the
fundamental ingredient of learning. 
 
This Lagrangian, however, does not convey 
a reasonable interpretation for a learning theory, since
one very much would like $\gamma>0$, so as $\gamma^{-1}$
could be nicely interpreted as a temporal regularization
parameter. Before exploring a different interpretation, we 
notice in passing that large values of $\theta/m$, which
corresponds with strong dissipation on small masses
yields the gradient flow
\[
	\dot{q} = - \frac{1}{\theta} V_{q}
\]

\section{Laws of learning and gradient flow}
While the discussion in the previous section provides
a somewhat surprising links with mechanics, the interpretation
of the learning as a problem of least actions is not very 
satisfactory since, just like in mechanics, we only
end up into stationary points of the actions that 
are typically saddle points. 

We will see that an appropriate choice of the Lagrangian
function yields truly laws of nature where Euler-Lagrange
equations turns out to minimize corresponding actions that
are appropriate to capture learning tasks.
We consider kinetic energies that also involve the
acceleration and two different cases which depend on the
choice of $h$.
The new action is
\begin{equation}
	{\cal A}_{2} = 
	\int_{0}^{T} dt \ L(t,q(t),\dot{q}(t),
	\ddot{q}(t)),
\end{equation}
where $L = h \bar{L}$.
In the continuum setting, the corresponding Euler-Lagrange equations can be determined by considering the variation
associated with $q \leadsto q+s v$, where $v$ is a variation
and $s \in \bbR$. We have
\begin{align}
	\delta {\cal A}_{2} = s \int_{0}^{T} dt \
	(L_{q} v + L_{p} \dot{v} + L_{a} \ddot{v}).
\end{align}
If we integrate by parts, we get
\begin{align*}
	&\int_{0}^{T} dt \  L_{p} \dot{v} = 
	- \int_{0}^{T} dt \ v \frac{d}{dt} L_{p} + \big[v L_{p} \big]_{0}^{T}\\
	&\int_{0}^{T} dt \  L_{a} \ddot{v} = 
	- \int_{0}^{T} dt \ \dot{v} \frac{d}{dt} L_{a} 
	+ \big[\dot{v} L_{a} \big]_{0}^{T} = 
	\int_{0}^{T} dt \ v \frac{d^{2}}{d t^{2}} L_{a}
	-  \bigg[v \frac{d}{dt}L_{a} \bigg]_{0}^{T}
	+ \big[\dot{v} L_{a} \big]_{0}^{T}, 
\end{align*}
and, therefore, the variation becomes
\begin{align*}
	\delta {\cal A}_{2} = s \int_{0}^{T} dt \ v
	\bigg(
		\frac{d^{2}}{d t^{2}} L_{a} -
		\frac{d}{dt} L_{p}
		+ L_{q}
	\bigg)
	+ \bigg[v \bigg(L_{p} - \frac{d}{dt} L_{a}\bigg)\bigg]_{0}^{T}
	+ \big[\dot{v} L_{a} \big]_{0}^{T}=0.
\end{align*}
Now, suppose we give the initial conditions on $q$ and $\dot{q}$. In that case
we can promptly see that this is equivalent with posing
$v(0)=0$ and $\dot{v}(0)=0$. Hence, we get the 
Euler-Lagrange equation when posing
\[
	v (T) \bigg(L_{p}\big |_{t=T} 
	- \frac{d}{dt} L_{a} \bigg|_{t=T} \bigg)
	+ \dot{v} (T) L_{a}  \big|_{t=T} =0.
\]
Now if we choose $v(t)$ as a constant we immediately get
\begin{equation}
	L_{p}\big |_{t=T} 
	- \frac{d}{dt} L_{a} \bigg|_{t=T}=0,
\label{BC1}
\end{equation}
while if we choose $v$ as an affine function, when considering
the above condition we get 
\begin{equation}
	L_{a} \bigg|_{t=T}=0.
\label{BC2}
\end{equation}
Finally, the stationary point of the action corresponds with
the Euler-Lagrange equations
\begin{align}
	\frac{d^{2}}{d t^{2}} L_{a} -\frac{d}{dt} L_{p} + L_{q} = 0,
\end{align}
that holds along with Cauchy initial conditions on $q(0), \dot{q}(0)$
and  boundary conditions~\ref{BC1} and~\ref{BC2}.

Now, let us consider the case in which $L = h \bar{L}$.
The Euler-Lagrange equations become
\begin{equation}
	\frac{d^{2}}{d t^{2}} \bar{L}_{a} + 2 \frac{\dot{h}}{h}
	\frac{d}{dt} \bar{L}_{a} + \frac{\ddot{h}}{h} \bar{L}_{a}
	- \frac{\dot{h}}{h} \bar{L}_{p} -\frac{d}{dt} \bar{L}_{p} + L_{q}=0.
\end{equation}
If we consider again the case $\bar{L} = T +\gamma V$ we
get 
\begin{equation}
	\frac{d^{2}}{d t^{2}} T_{a} + 2 \frac{\dot{h}}{h}
	\frac{d}{dt} T_{a} + \frac{\ddot{h}}{h} T_{a}
	- \frac{\dot{h}}{h} T_{p} -\frac{d}{dt} T_{p} 
	+ \gamma V_{q}=0.
\label{EL-2nd-order-T}
\end{equation}
Now we consider the kinetic energy associated with the
differential operator 
$P=\alpha_{1} \frac{d}{dt} + \alpha_{2} \frac{d^{2}}{d t^{2}}$
\begin{align}
\begin{split}
	T = \frac{1}{2 \theta^{2}}(P q)^{2} &= 
	\frac{1}{2\theta^{2}}(\alpha_{1} \dot{q}+\alpha_{2} \ddot{q})^{2}
	= \frac{1}{2} \frac{\alpha_{1}^{2}}{\theta^{2}} \dot{q}^{2}
	+ \frac{\alpha_{1} \alpha_{2}}{\theta^{2}} \dot{q} \ddot{q}
	+ \frac{1}{2}\frac{\alpha_{2}^{2}}{\theta^{2}} \ddot{q}^{2}\\
\end{split}
\label{Kinetic2ndOrder}
\end{align}
Let us consider the following two different cases of $h(t)$.
In both cases, they convey the unidirectional structure of time.
\begin{enumerate}
\item [$i$.] $h(t) = e^{\theta t}$\\
In this case,  when plugging the kinetic energy in 
Eq.~\ref{Kinetic2ndOrder} into  Eq.~\ref{EL-2nd-order-T}
we get
\begin{equation}
	\frac{1}{\theta^{2}}q^{(4)} + \frac{2}{\theta} q^{(3)} + 
	\frac{\alpha_{1} \alpha_{2} \theta 
	+ \alpha_{2}^{2} \theta^{2}-\alpha_{1}^{2}}
	{\alpha_{2}^{2} \theta^{2}}
	\ddot{q}
	+ \frac{\alpha_{1} \alpha_{2} \theta^{2} - \alpha_{1}^{2} \theta}{\alpha_{2}^{2}  \theta^{2}} \dot{q} + \frac{\gamma}{\alpha_{2}^{2}} V_{q} = 0.
\label{FourthOrderStab}
\end{equation}
These equations hold along with Cauchy conditions and boundary
conditions given by Eq.~\ref{BC1} and~\ref{BC2}, that turn out
to be
\begin{align}
	&\frac{\alpha_{1}^{2}}{\theta^{2}} \dot{q}(T) 
	 - \frac{\alpha_{2}^{2}}{\theta^{2}} q^{(3)}(T)  =0\\
	&\frac{\alpha_{1} \alpha_{2}}{\theta^{2}} \dot{q}(T) 
	+ \frac{\alpha_{2}^{2}}{\theta^{2}} \ddot{q}(T) = 0.
\end{align}
A possible satisfaction is 
$\dot{q}(0)=\ddot{q}(0)=q^{(3)}(0) = 0$.
Notice that as $\theta \rightarrow \infty$ the Euler-Lagrange
Eq.~\ref{FourthOrderStab} reduces to
\begin{equation}
	\ddot{q}
	+ \frac{\alpha_{1}}{\alpha_{2}} \dot{q} 
	+ \frac{\gamma}{\alpha_{2}^{2}} V_{q} = 0.
\end{equation}
and the corresponding boundary conditions are 
always verified.

\item [$ii.$] $h(t) = e^{-t/\epsilon}$ \\
Let us assume that $\beta = 0$ in the 
kinetic energy~\ref{Kinetic2ndOrder} and $h(t) = e^{-t/\epsilon}$. 
In particular we consider the action
\begin{equation}
	{\cal A} = \int_{0}^{T} dt \ e^{-t/\epsilon} \bigg(
		\frac{1}{2} \epsilon^{2} \rho \ddot{q}^{2}
		+\frac{1}{2} \epsilon \nu \dot{q}^{2}
		+V(q,t)
	\bigg)
\end{equation}
In this case the Lagrange equations turn out to be
\begin{equation}
	\epsilon^{2} \rho q^{(4)} - 2 \epsilon \rho q^{(3)}
	+ (\rho - \epsilon \nu) \ddot{q} + \nu \dot{q} + 
	\gamma V_{q} = 0,
\label{FourthOrderUns}
\end{equation}
along with the boundary conditions
\begin{align}
	&\epsilon^{2} \rho \ddot{q}(T)=0 \\
	&\epsilon \nu \ddot{q}(T) - \rho \epsilon^{2} q^{3}(T)=0.
\end{align}
Interesting, as $\epsilon \rightarrow 0$ the Euler-Lagrange 
equations become:
\begin{equation}
	\rho \ddot{q} + \nu \dot{q} + 
	\gamma V_{q} = 0,
\end{equation}
where the boundary conditions are always satisfied. 
%
%

%
%
\end{enumerate}
\begin{remark}
	Notice that while we can choose the parameters in 
	such a way that Eq.~\ref{FourthOrderStab} is stable,
	the same does not hold for Eq.~\ref{FourthOrderUns}.
	Interestingly, stability can be gained for $\epsilon=0$,
	which is corresponds with a singular solution.
	Basically if we denote by $q_{\epsilon}$ the solution
	associated with $\epsilon \in \bbR$, we have 
	that $q_{\epsilon}$ does not approximate $q$ corresponding
	at $\epsilon=0$ in case in which we can choose 
	arbitrarily large domains $[0,T]$. 
\end{remark}

%

\section{Conclusions}
 While machine learning is typically framed in the statistical
 setting, in this case time is exploited in such a way that
 one relies on a sort of underlying ergodic principle according
 to which statistical regularities can be captured in time. 
 This paper shows that the continuous nature of time
 gives rise to computational models of learning that
 can be interpreted as laws of nature. Unlike traditional
 stochastic gradient, the theory suggests that, just like
 in mechanics, learning is driven by the Euler-Lagrange
 equations that minimize a sort of functional risk.
 The collapsing from forth- to second-order differential equations
 opens the doors to an in-depth theoretical and experimental
 investigation.

\section*{Acknowledgments}
We thank Giovanni Bellettini for insightful discussions. 

\bibliography{nn,corr}
\bibliographystyle{plain}

\end{document}